\documentclass[conference]{IEEEtran}
\IEEEoverridecommandlockouts

\usepackage{url}
\usepackage{float}
\usepackage{cite}
\usepackage{amsmath,amssymb,amsfonts}
\usepackage{algorithmic}
\usepackage{graphicx}
\usepackage{textcomp}
\usepackage{xcolor}
\usepackage{mathtools}
\usepackage{bm} 

\def\BibTeX{{\rm B\kern-.05em{\sc i\kern-.025em b}\kern-.08em
    T\kern-.1667em\lower.7ex\hbox{E}\kern-.125emX}}

\begin{document}

\title{Deep Learning-Based Lunar Crater Terrain Relative Navigation}

\author{\IEEEauthorblockN{Batu Candan}
\IEEEauthorblockA{\textit{Aerospace Engineering Department} \\
\textit{Iowa State University}\\
Ames, Iowa, USA \\
dukynuke@iastate.edu}
\and
\IEEEauthorblockN{Simone Servadio}
\IEEEauthorblockA{\textit{Aerospace Engineering Department} \\
\textit{Iowa State University}\\
Ames, Iowa, USA \\
servadio@iastate.edu}
}

\maketitle

\begin{abstract}
Accurate position estimation is crucial for the successful implementation of future lunar landings using autonomous vehicles, especially in dangerous environments with sparse terrain features. In this paper, we propose a terrain relative navigation (TRN) algorithm combining our deep-learning crater detector, which was designed specifically for the NASA Crater Detection Challenge problem, and an Extended Kalman Filter (EKF). Our detector analyzes crater features from the monocular images acquired from orbit, and their matches with craters from a global database are identified via a Hungarian assignment approach followed by the consensus-based outliers removal method. The estimated measurements are then used to refine an EKF, where spacecraft pose estimation in the Lunar-Centered Lunar-Fixed (LCLF) frame of reference, augmented with altitude aiding information, constrains radial drift. The simulation results indicate that even if the spacecraft is off from its actual location up to 5 km, TRN could recover from this situation, achieving navigation error reduction to a few hundred meters. It should be noted that in order to maintain crater feature correspondences, it is important to match the image resolution and the scales within the scene to the detector training set distribution.
\end{abstract}


\section{Introduction}
When missions aim for tough Moon landing spots like the south pole, counting only on a standard inertial navigation system (INS) falls short. Over long stretches without contact, small mistakes add up until they’re too big to ignore \cite{b1, b3, serva1, serva2}. That is where TRN steps in - using outside sensors to match what the craft sees with stored maps, keeping position fixes accurate. Instead of just moving forward blindly, the system checks itself against real landscape clues. Craters work well for this job on the Moon: there are plenty, they stay put across years, plus scientists already mapped them from orbit \cite{downesSM}. Yet spotting these craters reliably isn’t easy. Shadows shift, angles change, and surfaces look different depending on how light hits, making consistent matches tricky. Most older ways of finding craters - using edge tracing, shape guessing, or fixed templates - tend to slow down fast and react poorly when lighting or surface details shift suddenly. Lately, systems built around deep neural networks, especially those shaped like layered filters, have handled such jobs better on planet photos, pulling out shapes more reliably than standard tools even when views get messy or unclear \cite{ieee2, chand2, silv, ieee3}.

Motivated by these developments, this work builds upon a custom crater detection network originally developed in the context of the NASA Crater Detection Challenge and adapts it for use within a crater-based terrain relative navigation pipeline. The detector is implemented using a U-Net++-based segmentation architecture \cite{unet} and is used to extract crater-related image features from monocular orbital imagery. These detections are then associated with entries from a global lunar crater database and incorporated into an EKF that estimates spacecraft position and velocity in the LCLF frame. In this way, the proposed framework combines a learned visual front-end with a recursive state estimator, allowing crater observations to be used directly in the navigation loop. To evaluate the integrated system, synthetic lunar imagery is generated from the Lunar Reconnaissance Orbiter Camera (LROC) Wide Angle Camera (WAC) global mosaic \cite{lroc}. So, it provides a controlled environment for studying the interaction between crater detection quality, feature association, and EKF-based state estimation under the assumed imaging conditions. Rather than focusing only on crater detection as a standalone computer vision problem, the present work examines how the detector behaves when embedded inside a full navigation architecture, where false associations, missed detections, and measurement geometry directly affect filter performance.

The remainder of this paper is organized as follows. Section II introduces the coordinate frame definition, the crater detection framework, and the synthetic data generation process. Section III presents the feature association procedure and the EKF formulation. Section IV describes the simulation setup and summarizes the navigation results. Finally, Section V provides concluding remarks and directions for future work.

\section{Preliminaries}

\subsection{Coordinate Frames}
\begin{figure}[H]
    \centering
    \includegraphics[width=\linewidth]{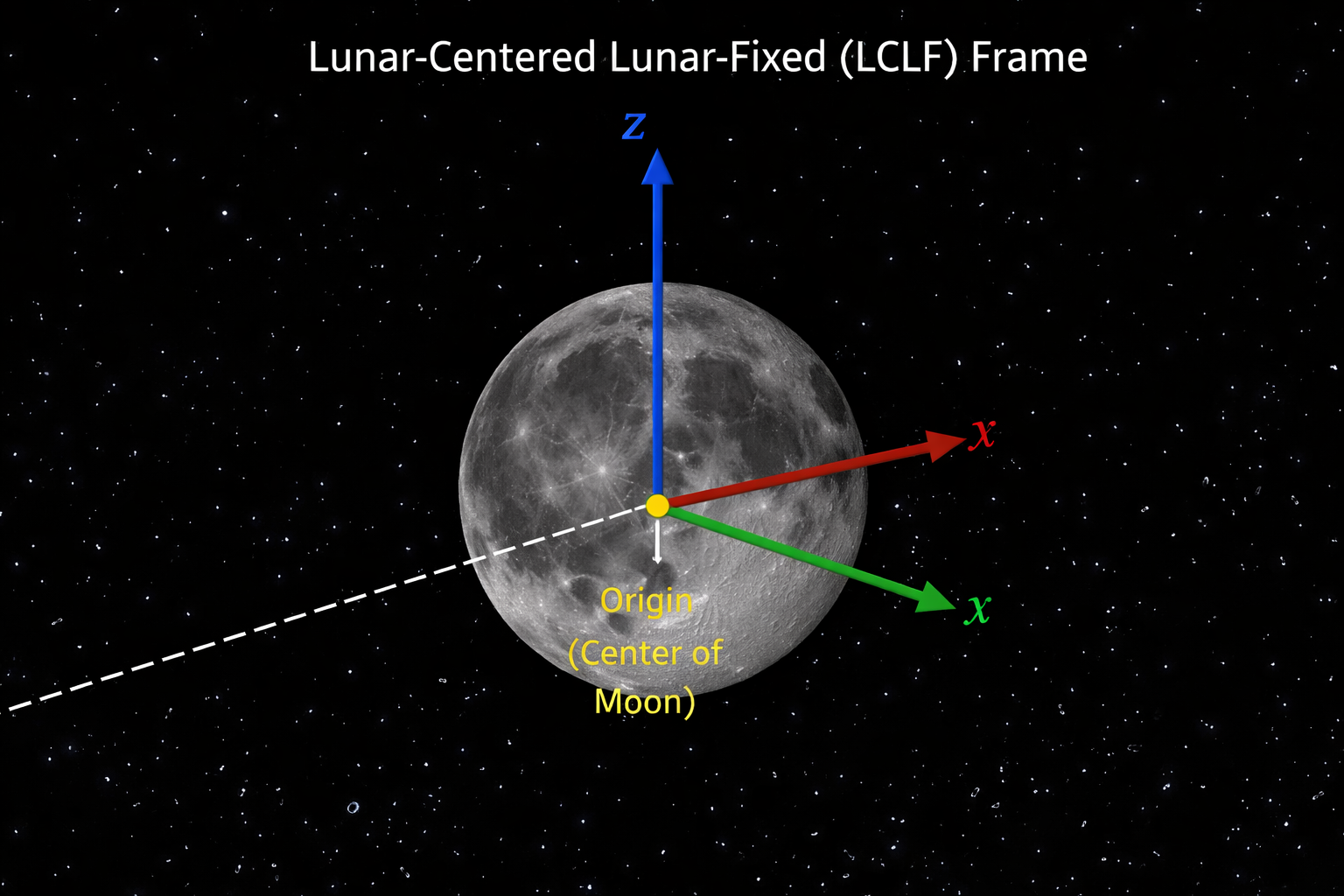}
    \caption{Illustration of the Lunar-Centered Lunar-Fixed (LCLF) coordinate frame used in this work}
    \label{fig:lclf_frame}
\end{figure}

The navigation state is estimated in the Lunar-Centered Lunar-Fixed (LCLF) frame. The spacecraft position $\mathbf{r} \in \mathbb{R}^3$ and velocity $\mathbf{v} \in \mathbb{R}^3$ constitute the state vector $\mathbf{x} = [\mathbf{v}^\top, \mathbf{r}^\top]^\top$. Figure~\ref{fig:lclf_frame} illustrates the LCLF coordinate frame adopted in this study. In this frame, the origin is placed at the center of the Moon, and the spacecraft state is represented by its position and velocity vectors relative to this body-fixed reference system.

\subsection{Detector Development and NASA Challenge Background}
The crater detection component used in this work builds upon earlier development carried out in the context of the NASA Crater Detection Challenge \cite{nasaChallenge}. The challenge focused on the reliable identification of lunar crater rims and ellipse-based characterization of visible craters from orbital imagery, which closely aligns with the needs of crater-based terrain relative navigation. In that earlier effort, a deep learning-based crater segmentation pipeline was developed to identify crater-like structures from planetary surface imagery and to provide robust feature localization under varying surface appearance conditions. The corresponding author participated in the NASA Crater Detection Challenge and placed 8th on the public Topcoder leaderboard \cite{topcoderChallenge}. This experience provided the initial motivation and practical foundation for the visual front-end adopted in this study. The original detector developed for the NASA Crater Detection Challenge was trained using the challenge-provided image set together with ellipse-based crater annotations stored in a ground-truth CSV file. The dataset contains $4{,}150$ unique images, which were split at the image level into $3{,}528$ training images and $622$ validation images using a $15\%$ hold-out validation partition. For each image, the ground-truth labels provide the crater ellipse center coordinates, semimajor axis, semiminor axis, and rotation angle. Before target generation, crater annotations were filtered according to the challenge rules. These filtering rules ensured that only valid, fully observable crater targets were used for supervision.

The challenge-oriented detector was trained on RGB images resized to $512 \times 512$ pixels. Two supervision targets were generated for each image: an edge mask representing the crater rim and a center heatmap representing crater centroids. The rim target was constructed by rasterizing the annotated ellipse boundary, while the center target was generated as a Gaussian heatmap centered on each valid crater. This yielded a two-channel training target that jointly encoded crater geometry and crater location. Data augmentation was used to improve robustness across varying viewing and illumination conditions. The augmentation pipeline included random horizontal and vertical flips, random $90^\circ$ rotations, adaptive histogram equalization (CLAHE), random gamma or brightness--contrast perturbations, and additive Gaussian noise, followed by image normalization using ImageNet statistics. The network was implemented as a U-Net++ model with a ResNet50 encoder initialized with ImageNet weights. The detector was trained using a custom dual-task loss composed of a binary cross-entropy term for crater rim prediction and a Smooth L1 term for crater center heatmap prediction, with the two terms weighted to emphasize both boundary extraction and centroid localization.

\begin{figure}
    \centering
    \includegraphics[width=\linewidth]{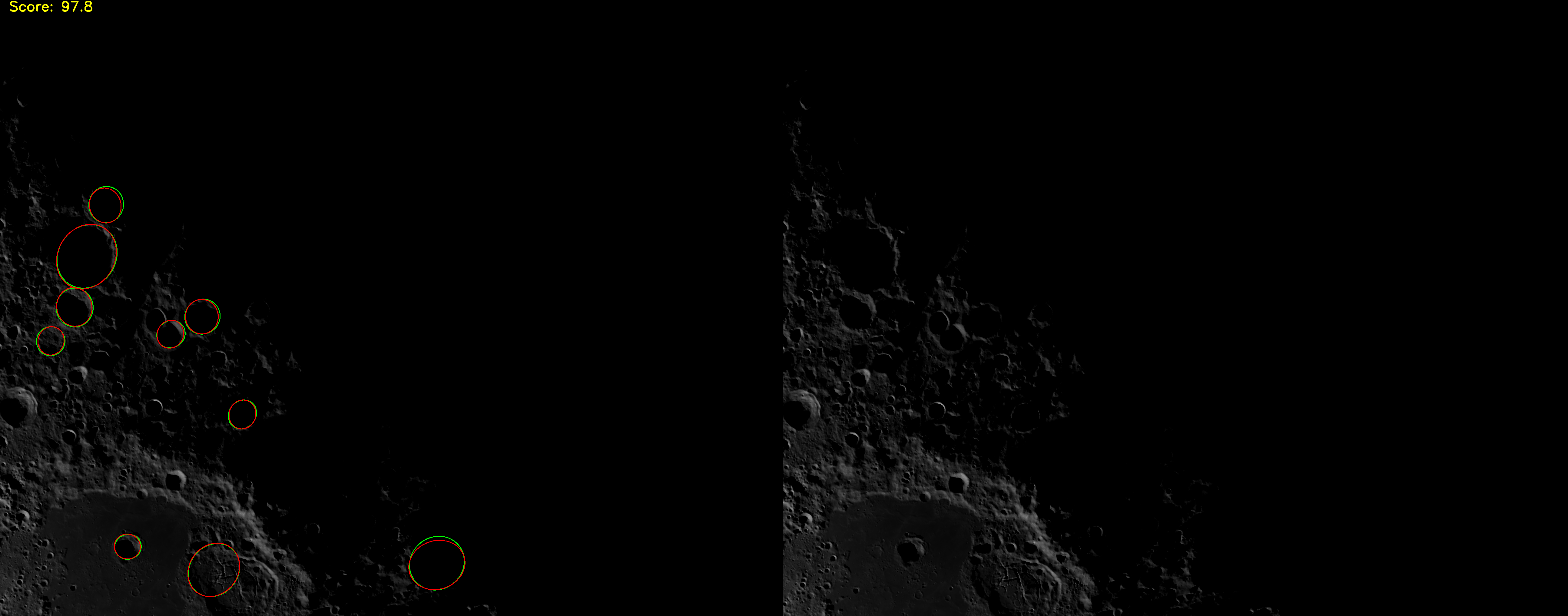}
    \caption{Representative qualitative result from the NASA Crater Detection Challenge under relatively bright illumination conditions}
    \label{fig:sim2}
\end{figure}

\begin{figure}
    \centering
    \includegraphics[width=\linewidth]{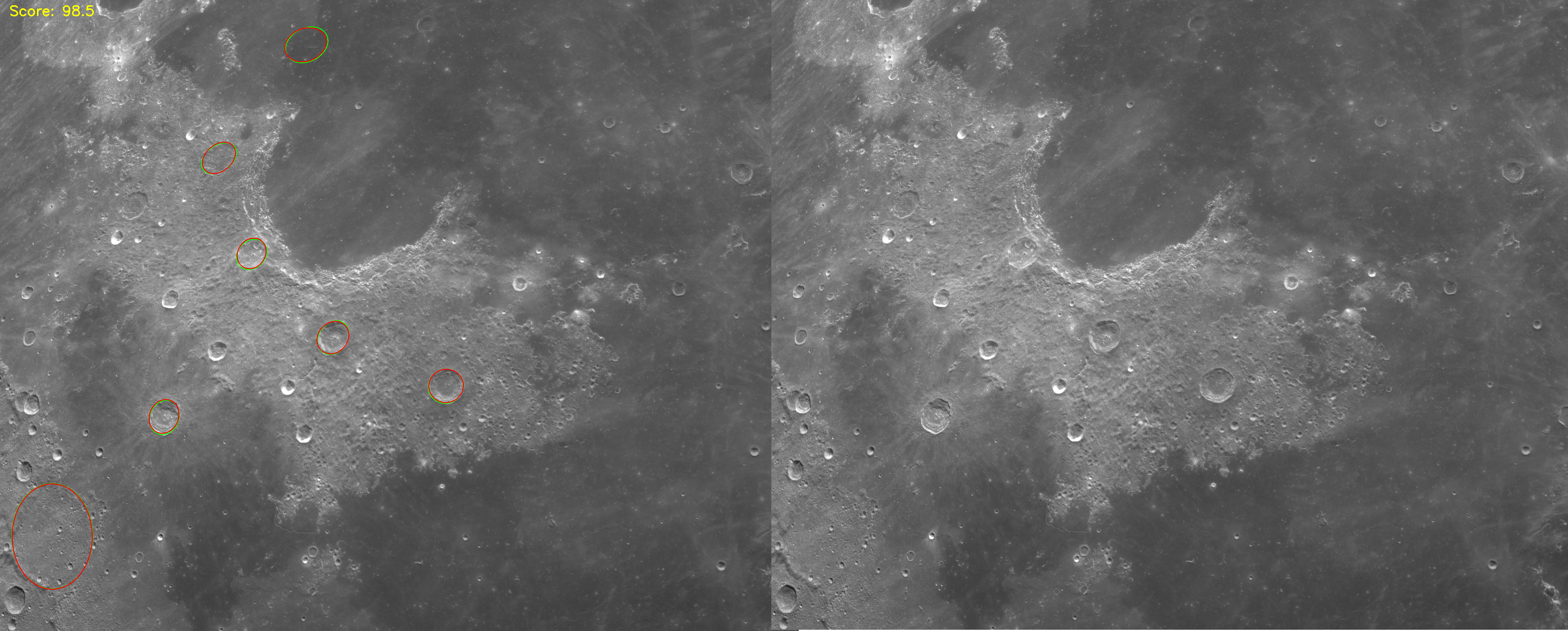}
    \caption{Representative qualitative result from the NASA Crater Detection Challenge under low-illumination and shadowed conditions}
    \label{fig:sim3}
\end{figure}

Optimization was performed using AdamW with a learning rate of $10^{-4}$ and weight decay of $10^{-5}$. A learning-rate scheduler based on validation loss reduction was used during training, and early stopping was applied with a patience of $15$ epochs. Mixed-precision training was employed to improve computational efficiency, and the best-performing model was selected according to the minimum validation loss. Figures~\ref{fig:sim2} and \ref{fig:sim3} show representative qualitative results from the earlier challenge-oriented detector under two different illumination conditions (on the left, the resulting images are shown, and on the right, the raw images are shown). In both examples, green ellipses denote ground-truth crater annotations, while red ellipses denote the predicted crater detections. The close agreement between the predicted and reference ellipses indicates that the detector accurately localizes the annotated craters in these representative cases, while preserving both the crater position and the apparent rim geometry. These results motivated the use of the challenge-derived detector as the basis for the visual front-end that was later adapted to the lunar terrain relative navigation framework presented in this paper.

The transition from the challenge-oriented detector to the present navigation framework required several modifications. First, the detector was retrained and evaluated on the custom lunar dataset generated in this work, rather than being used only in its original challenge setting. Second, the supervision format was organized to support both crater rim and crater center extraction, allowing the downstream processing stage to estimate crater location and apparent crater size in the image plane. Third, the detector outputs were integrated with a catalog-based feature matching stage and an EKF, enabling crater observations to be used directly as navigation measurements. This challenge heritage is important because it provides the basis for the learned visual front-end used in this paper, while the primary contribution of the present work lies in embedding that detector within a crater-based terrain relative navigation architecture for lunar state estimation.

\subsection{Crater Detection via U-Net++}
A custom crater detection network is used as the visual front-end of the proposed terrain relative navigation framework. The model is based on a U-Net++ architecture with a ResNet50 encoder pre-trained on ImageNet \cite{unet}. Unlike a single-output crater segmentation formulation, the network is trained to predict two complementary binary masks: a crater rim mask and a crater center mask. This two-channel representation improves the robustness of the downstream feature extraction stage by jointly encoding crater extent and crater centroid information. The input to the network is a $256 \times 256$ RGB image. Each image is normalized channel-wise using the ImageNet mean and standard deviation values before being passed to the network. The corresponding training label is a two-channel binary target, where the first channel represents crater rims and the second channel represents crater centers. 

During training, the crater rim and center masks are stored in a single image container and separated into individual channels before conversion to tensor format. The network is trained using the Adam optimizer with a learning rate of $10^{-4}$, batch size of $8$, and $180$ training epochs. A hold-out validation split of $10\%$ of the available samples is used for model selection. The loss function is defined as an equally weighted combination of binary cross-entropy with logits and Dice loss, which provides a balance between pixel-wise classification accuracy and region-level overlap. The best-performing model is selected according to the lowest validation loss and saved for subsequent inference. During inference, the network outputs two probability maps corresponding to crater rims and crater centers. A sigmoid activation is applied to the logits, and fixed thresholds are used to generate binary masks for both channels. Connected components are extracted from the crater center mask to identify candidate crater centroids. These centroids are then associated with nearby contours extracted from the crater rim mask, and ellipses are fitted to the selected contours to estimate crater center coordinates and apparent crater size in the image plane. The resulting crater detections are represented by image-plane center coordinates and ellipse dimensions, which are then passed to the feature association and navigation filter stages.

\subsection{Data Generation}
The dataset used in this work is generated from the LROC WAC global mosaic, stored as a georeferenced GeoTIFF map \cite{lroc}. A custom image generation pipeline is used to extract localized orthographic views of the lunar surface centered at specified latitude and longitude coordinates. For each sample, a square $256 \times 256$ image is created to represent a fixed-width surface patch spanning $100$ km. The image intensities are obtained by remapping the GeoTIFF data onto the orthographic grid and applying percentile-based contrast normalization. To generate crater labels, a global lunar crater catalog \cite{data1, data2, data3} is filtered using the local field of view around the selected image center. Crater centers are projected into the orthographic image plane, and only visible craters above a minimum diameter threshold are retained. For each valid crater, a circular rim annotation is drawn to define the crater boundary target, and a small filled disk is drawn at the crater center to define the centroid target. These two annotations form the two-channel supervision used for network training.

Two datasets are generated for this study. The training set consists of randomly sampled image locations across the lunar surface within a prescribed latitude range, producing diverse crater appearances and local terrain conditions. The test set is generated along a deterministic sequential trajectory intended for navigation filter evaluation. In the test sequence, the spacecraft is assumed to move along a simplified orbital path at a fixed altitude of $100$ km with constant sampling interval. For each test frame, the corresponding image, crater mask, and metadata, including latitude, longitude, altitude, and time step, are stored. This metadata is later used to initialize and evaluate the EKF.

\section{Methodology}

\subsection{Feature Matching and Data Association}
Unlike simultaneous localization and mapping (SLAM) frameworks that dynamically augment the state vector with mapped landmarks \cite{downesSM}, or optical flow methods that track unknown features frame-to-frame \cite{b3}, the proposed architecture maintains a strictly 6-dimensional state vector composed only of the spacecraft's position and velocity. Crater detections are treated strictly as absolute line-of-sight measurements to known catalog features, bounded by the CNN's characterized pixel variance. This absolute navigation formulation ensures the filter remains computationally lightweight and immune to the compounding drift associated with frame-to-frame feature tracking.

\begin{figure}
    \centering
    \includegraphics[width=\linewidth]{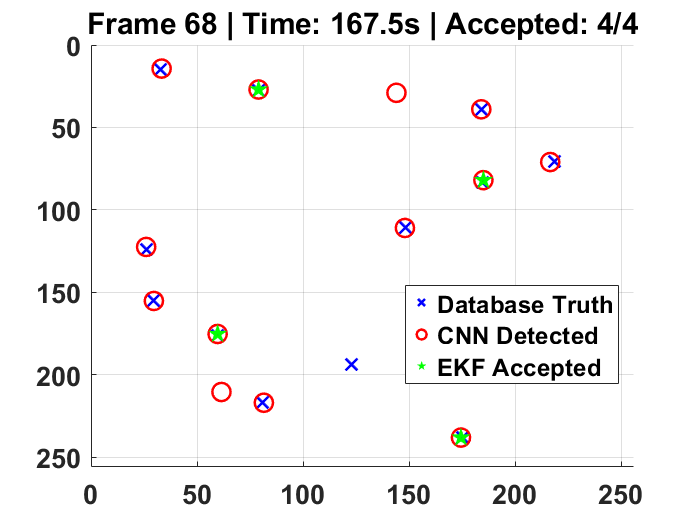}
    \caption{Data association and quadrant spread logic in the image plane. Red circles indicate raw CNN detections, blue crosses denote projected database truth, and green stars highlight the four spatially distributed fiducial markers successfully matched and accepted by the EKF for the measurement update.}
    \label{fig:quadrant_spread}
\end{figure}

To associate the CNN detections with the lunar crater database, a robust quadrant spread algorithm is employed. Rather than processing all visible craters, which often cluster locally and degrade the Geometric Dilution of Precision (GDOP), the algorithm isolates up to four spatially distributed fiducial markers per frame. Four virtual anchor points are defined at the quadrants of the $256 \times 256$ image plane (with a $25\%$ margin from the edges). For each anchor, the CNN detections are sorted in $O(N \log N)$ time based on their Euclidean distance to the anchor, and the closest unique detection is selected. This spatial gating guarantees that the EKF measurement updates are driven by features widely spread across the camera's field of view. Figure~\ref{fig:quadrant_spread} illustrates this robust association strategy, highlighting the spatial distribution of the four selected fiducial markers among the full set of raw detections.

Once the spatially distributed CNN detections are isolated, they are matched to the projected catalog craters using a nearest-neighbor search. Let $\mathbf{u}_{det}$ be the image-plane coordinates of a selected CNN detection, and $\mathbf{u}_{proj}$ be the projected center of a catalog crater based on the a priori state estimate. A match is accepted if the Euclidean pixel distance falls below a strict association threshold:
\begin{equation}
\|\mathbf{u}_{det} - \mathbf{u}_{proj}\| < \tau_{match}
\end{equation}
where $\tau_{match} = 3.0$ pixels in the current implementation. This computationally efficient quadrant sorting and thresholding strategy avoids the high overhead of global optimization methods, such as the Hungarian algorithm, while the spatial spreading mathematically ensures an optimal geometric baseline for the line-of-sight measurement updates.

\subsection{Extended Kalman Filter Design}
The navigation state is estimated using a discrete-time EKF. Let the state vector be defined as,
\begin{equation}
\mathbf{x}_k =
\begin{bmatrix}
\mathbf{v}_k \\
\mathbf{r}_k
\end{bmatrix}
=
\begin{bmatrix}
v_x & v_y & v_z & x & y & z
\end{bmatrix}^\top,
\end{equation}
where $\mathbf{v}_k \in \mathbb{R}^3$ and $\mathbf{r}_k \in \mathbb{R}^3$ denote the spacecraft velocity and position, respectively, in the LCLF frame. At each time step, the EKF prediction stage is written as
\begin{align}
\hat{\mathbf{x}}_{k|k-1} &= f\!\left(\hat{\mathbf{x}}_{k-1|k-1}\right), \\
\mathbf{P}_{k|k-1} &= \mathbf{F}_{k-1}\mathbf{P}_{k-1|k-1}\mathbf{F}_{k-1}^\top + \mathbf{Q}_{k-1},
\end{align}
where $f(\cdot)$ is the nonlinear process model, $\mathbf{F}_{k-1}$ is the corresponding state transition Jacobian, and $\mathbf{Q}_{k-1}$ is the process noise covariance. When crater measurements are available, the correction step is given by
\begin{align}
\mathbf{y}_k &= \mathbf{z}_k - h\!\left(\hat{\mathbf{x}}_{k|k-1}\right), \\
\mathbf{S}_k &= \mathbf{H}_k \mathbf{P}_{k|k-1} \mathbf{H}_k^\top + \mathbf{R}_k, \\
\mathbf{K}_k &= \mathbf{P}_{k|k-1}\mathbf{H}_k^\top \mathbf{S}_k^{-1}, \\
\hat{\mathbf{x}}_{k|k} &= \hat{\mathbf{x}}_{k|k-1} + \mathbf{K}_k \mathbf{y}_k, \\
\mathbf{P}_{k|k} &= (\mathbf{I} - \mathbf{K}_k \mathbf{H}_k)\mathbf{P}_{k|k-1}(\mathbf{I} - \mathbf{K}_k \mathbf{H}_k)^\top
+ \mathbf{K}_k \mathbf{R}_k \mathbf{K}_k^\top,
\end{align}
where $\mathbf{y}_k$ is the innovation, $\mathbf{H}_k$ is the measurement Jacobian, $\mathbf{R}_k$ is the measurement noise covariance, and $\mathbf{K}_k$ is the Kalman gain.

For state propagation, the filter uses a discrete-time kinematic model with gravitational acceleration evaluated from the current position estimate. The propagation equations are
\begin{align}
\mathbf{v}_k &= \mathbf{v}_{k-1} + \mathbf{a}_{k-1}\Delta t, \\
\mathbf{r}_k &= \mathbf{r}_{k-1} + \mathbf{v}_{k-1}\Delta t + \frac{1}{2}\mathbf{a}_{k-1}\Delta t^2,
\end{align}
where the acceleration is modeled as
\begin{equation}
\mathbf{a}_{k-1} = -\mu_m \frac{\mathbf{r}_{k-1}}{\|\mathbf{r}_{k-1}\|^3},
\end{equation}
and $\mu_m$ denotes the lunar gravitational parameter. The measurement model is formed from matched crater landmarks projected into the image plane. For the $j$-th matched crater, the measurement equation is written as
\begin{equation}
\mathbf{z}_{j,k} = h(\mathbf{x}_k,\mathbf{p}_j) + \bm{\nu}_{j,k},
\end{equation}
where $\mathbf{p}_j$ denotes the cataloged crater location, $h(\cdot)$ is the image-plane projection model, and $\bm{\nu}_{j,k}$ is zero-mean measurement noise. In the present implementation, crater measurements are represented in pixel coordinates, and the measurement Jacobian is computed numerically with respect to the spacecraft position. Residual gating is applied in the image plane to reject inconsistent crater correspondences before the EKF correction step.

\section{Simulation and Results}

\subsection{Simulation Setup}
The proposed navigation framework was evaluated using a synthetic lunar image sequence generated from the test portion of the custom dataset described earlier. Each frame corresponds to a $256 \times 256$ orthographic image patch extracted from the lunar surface, with a fixed surface coverage of $100$ km. The reference imagery was derived from the LROC WAC global mosaic, and the associated crater annotations and trajectory metadata were used to support both crater-based measurement generation and navigation error evaluation. The navigation filter was tested on a simulated orbital sequence at a fixed altitude of $100$ km above the lunar surface. The test trajectory was sampled at uniform time intervals, and the corresponding metadata for each frame included the reference latitude, longitude, altitude, and sampling time step. For crater detection, the trained U-Net++ based network was applied to each image frame, and the resulting crater center and rim predictions were converted into ellipse-based crater features for downstream feature association.

\begin{figure}[H]
    \centering
    \includegraphics[width=\linewidth]{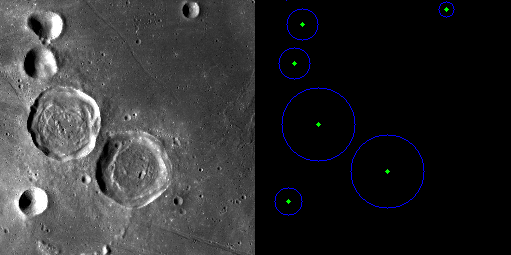}
    \caption{Example synthetic lunar image patch and its corresponding crater annotation mask}
    \label{fig:1}
\end{figure}

To assess robustness to initialization uncertainty, the EKF was evaluated in a Monte Carlo setting with $50$ independent trials. In each trial, the initial position estimate was perturbed by a fixed-magnitude $5$ km error with a randomized 3D direction, while the initial velocity estimate was perturbed by zero-mean Gaussian noise with a standard deviation of $10$ m/s on each Cartesian component. The initial covariance matrix was selected to reflect these assumed position and velocity uncertainties. In addition, the auxiliary altitude measurement used to constrain radial drift was corrupted by zero-mean Gaussian noise with a standard deviation of $1$ m. For the feature association stage, detected crater features were matched against projected catalog craters using a linear assignment procedure based on image-plane position and apparent crater diameter. A relaxed matching threshold was used during the first several frames to improve tolerance to the initial state error, after which a tighter threshold was applied. Candidate matches were then filtered using a consensus-based translation check to reject inconsistent correspondences before the EKF measurement update. The measurement noise for crater observations was modeled in the image plane using a fixed pixel-level covariance. Figure~\ref{fig:1} illustrates an example of a synthetic lunar image patch together with its corresponding crater annotation mask used in the training pipeline. The simulation outputs were analyzed in terms of total position error, horizontal surface error, and altitude error across the image sequence, and Monte Carlo statistics were used to evaluate the mean behavior and dispersion of the filter performance.

\subsection{Qualitative Crater Detection and Association}
\begin{figure}[H]
    \centering
    \includegraphics[width=0.9\linewidth]{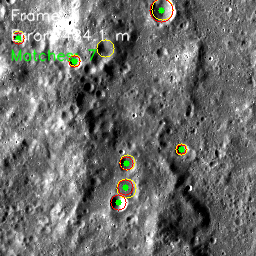}
    \caption{Example of crater detection and feature association on a simulated lunar image}
    \label{fig:2}
\end{figure}

To illustrate the measurement generation process used by the navigation filter, Fig.~\ref{fig:2} shows a representative frame from the simulated image sequence with the crater detection and association results overlaid. In each frame, the trained U-Net++-based detector produces crater candidates that are converted into ellipse-based image features. These detections are then compared against the catalog crater projections obtained from the current EKF state estimate. A linear assignment step is used to form candidate correspondences, followed by a consensus-based outlier rejection stage to retain only geometrically consistent matches.

In the visualization, the red contours indicate crater ellipses extracted from the image by the detector, while the yellow contours represent the projected crater locations predicted from the current navigation estimate. Green markers denote crater centers that were accepted as valid matches and used in the EKF measurement update. This qualitative result demonstrates that the proposed front-end is able to produce usable crater correspondences directly from the simulated orbital imagery, providing the feature measurements required for recursive state correction.

\subsection{Quantitative Sensor Characterization}
To rigorously evaluate the performance of the U-Net++ front-end, a large-scale global characterization was conducted, mirroring the validation methodology used in prior state-of-the-art benchmarks such as LunaNet \cite{downesSM}. A massive evaluation set of $6{,}500$ orthographic images was generated across the entire lunar sphere, employing uniform spherical sampling to prevent artificial clustering at the poles. The network processed these images, and the resulting detections were associated with the ground-truth catalog using a tight 2.0-pixel matching threshold.

\begin{figure}[H]
    \centering
    \includegraphics[width=\linewidth]{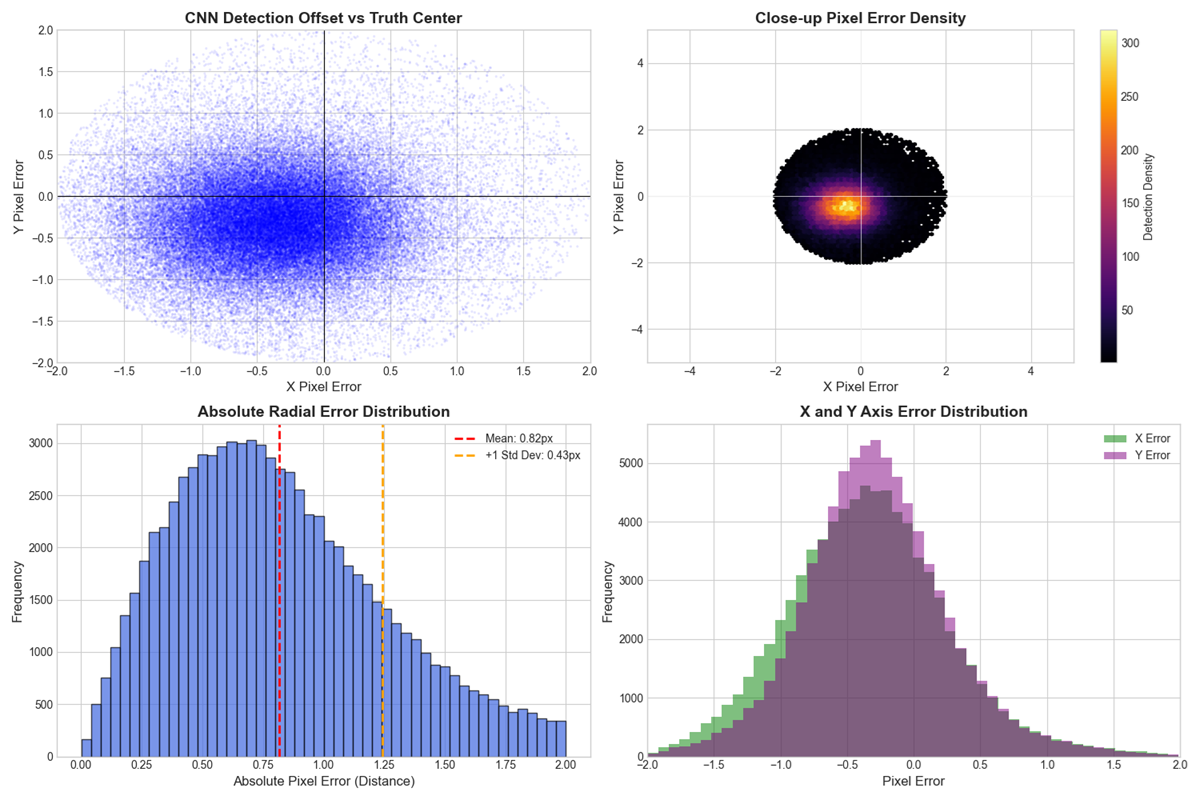}
    \caption{Sensor characterization across $6{,}500$ global lunar images ($80{,}293$ valid matches). Top left: CNN detection offset versus truth center. Top right: Close-up pixel error density hexbin. Bottom: Absolute radial and X/Y component error distributions.}
    \label{fig:characterization}
\end{figure}

This global sweep yielded $80{,}293$ valid crater matches. Statistical analysis of the image-plane residuals revealed a mean radial pixel error of $0.8163$ pixels and a standard deviation of $0.4285$ pixels. Notably, the X and Y component errors were tightly centered around zero, demonstrating that the network possesses no significant directional bias. Figure~\ref{fig:characterization} illustrates the error distribution across the matched dataset. Compared to previous CNN-based TRN pipelines, which reported standard deviations on the order of 1.4 pixels \cite{downesSM}, the proposed U-Net++ architecture achieves a fundamentally tighter uncertainty bound.

\subsection{Convergence Analysis}
Following the quantitative characterization of the U-Net++ front-end, the simulated camera model in the EKF was driven purely by the network's calibrated standard deviation of 0.4285 pixels. To guarantee bounded performance and prevent filter divergence under large initial errors, the base measurement covariance matrix $\mathbf{R}_k$ was established using the conservative $2\sigma$ bound of the characterized data. Additionally, an adaptive measurement covariance scaling term was introduced during the Kalman gain calculation to robustly handle transient non-linearities and unmodeled measurement biases.

\begin{figure}[H]
    \centering
    \includegraphics[width=0.95\linewidth]{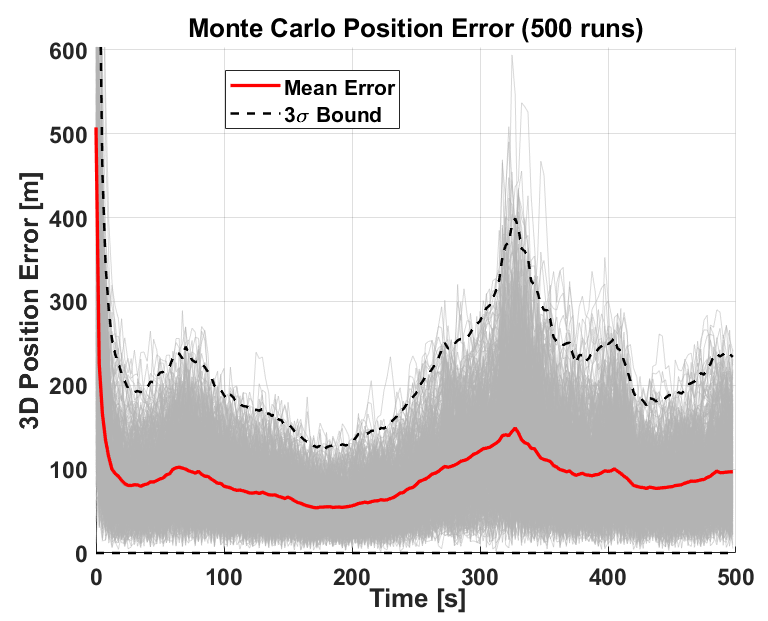}
    \caption{Monte Carlo 3D position error ($N=500$ runs).}
    \label{fig:pos_err}
\end{figure}

The navigation filter demonstrates rapid and stable convergence. To rigorously stress-test the system, the EKF was initialized with a 1 km standard deviation uncertainty per axis (yielding a highly dispersed initial 3D error volume). As shown in Fig.~\ref{fig:pos_err}, the filter successfully recovers from this extreme initial dispersion, aggressively reducing the mean 3D position error to a steady-state floor of approximately 50 to 100 meters. The $3\sigma$ confidence bounds remain strictly bounded throughout the 500-second descent sequence, successfully absorbing dynamic trajectory variations.

\begin{figure}[H]
    \centering
    \includegraphics[width=0.95\linewidth]{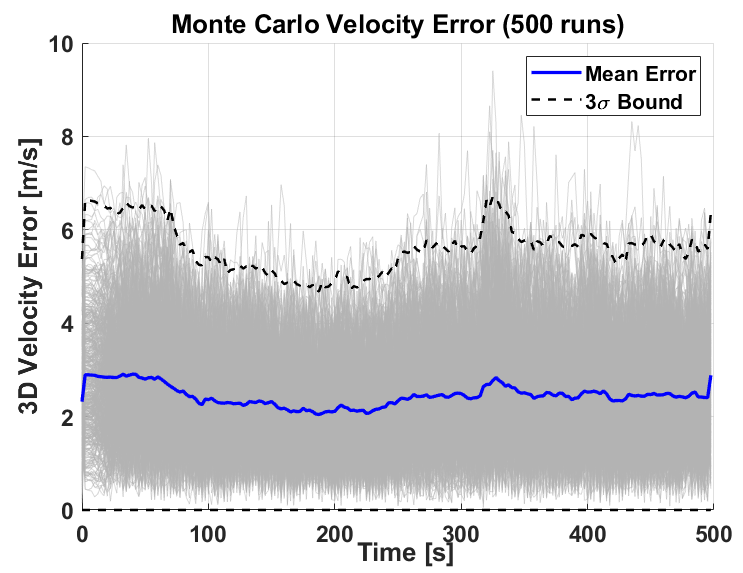}
    \caption{Monte Carlo 3D velocity error ($N=500$ runs).}
    \label{fig:vel_err}
\end{figure}

Fig.~\ref{fig:vel_err} illustrates the corresponding velocity estimation performance, where the mean 3D velocity error rapidly stabilizes near 2.5 m/s. It is critical to note that this convergence is achieved in a strictly autonomous, bearing-only configuration without the aid of a radar altimeter. By coupling the superior sub-pixel resolving power of the U-Net++ detector with a robust quadrant-spread data association strategy, the proposed formulation successfully tracks and bounds the spacecraft state. This demonstrates a clear performance margin over the baseline LunaNet architecture \cite{downesSM}, which relied on significantly smaller initial uncertainties to maintain track.

\section{Conclusion}
This paper presented a crater-based TRN framework for lunar applications that combines a custom deep learning-based crater detection network with an EKF for recursive state estimation. A core contribution of this work is the rigorous global characterization of the U-Net++ visual front-end, which was proven to achieve sub-pixel measurement variance across a highly diverse operational envelope, directly outperforming prior state-of-the-art CNN baselines. The proposed approach leverages this precision to extract crater features from monocular orbital imagery and associates them with entries from a global lunar crater database using a computationally efficient, robust quadrant-spread assignment procedure. 

Crucially, the resulting measurements are incorporated into an absolute navigation filter that maintains a strictly 6-dimensional state vector in the LCLF frame. Unlike SLAM frameworks that continuously inflate the state vector and covariance matrices with newly mapped landmarks, or optical flow methods prone to frame-to-frame drift, this absolute formulation relies purely on global catalog matching. This helps keep the architecture lightweight, bounded, and well-conditioned for the computational limits of future flight hardware. 

The simulation presented was trained using a set of synthetic lunar images created using the LROC WAC global mosaic. The use of these synthetic images allowed an end-to-end test of the crater detection, optimized geometric data association, and filter performance within a single pipeline. The simulations prove that the proposed bearing-only approach can recover from kilometer-level initial state uncertainty and rapidly down-track the navigation error to sub-100 m of navigation error without the use of a radar altimeter. 

In addition to the numerical results, this work clearly illustrates the benefit of making the detector training distribution congruent with the runtime image generation process, including, among others, the image scale, the surface coverage, and the density of features present. It also emphasizes the benefit of fusing a carefully trained vision front-end with model-based state estimation. Future work will take several possible directions in order to improve the current formulation. First, a more sensor representative configuration, including proper IMU characterization and dynamic illumination, can be introduced. Second, the features matching stage can be augmented through the use of an uncertainty driven matching method. Finally, the current scheme will be generalized from orbital flight toward the more challenging six-DOF landing problem, including dynamically varying viewing geometry, coupled translation and rotation, and dynamic hardware-in-the-loop limitations.

\section*{Acknowledgment}
This work was sponsored in part by the U.S. Air Force Office of Scientific Research (AFOSR) under grant number
FA9550-23-S-0001.

\vspace{12pt}

\end{document}